\documentclass[letterpaper, 10 pt, conference]{ieeeconf}  

\IEEEoverridecommandlockouts                              

\usepackage{graphicx} 
\usepackage{mathptmx} 
\usepackage{amsmath} 
\usepackage{amssymb}  

\usepackage{balance}
\usepackage{comment}

\usepackage{subcaption}
\usepackage{algorithm, algpseudocode}
\usepackage{siunitx}
\usepackage{booktabs}
\usepackage[table]{xcolor}
\usepackage{multirow}
\usepackage{makecell}

\title{\LARGE \bf
 Reference-Free Sampling-Based Model Predictive Control
}

\author{Fabian Schramm$^{1}$, Pierre Fabre$^{1}$, Nicolas Perrin-Gilbert$^{2}$ and Justin Carpentier$^{1}$
\thanks{$^{1}$Inria - Département d’Informatique de l’École normale supérieure, PSL Research University
        {\tt\small firstname.lastname@inria.fr}}%
\thanks{$^{2}$Sorbonne Université, CNRS, Institut des Systèmes Intelligents et de Robotique, ISIR}%
}

\begin{document}

\maketitle
\thispagestyle{empty}
\pagestyle{empty}

\begin{abstract}
We present a sampling-based model predictive control (MPC) framework that enables emergent locomotion without relying on handcrafted gait patterns or predefined contact sequences. Our method discovers diverse motion patterns, ranging from trotting to galloping, robust standing policies, jumping, and handstand balancing, purely through the optimization of high-level objectives. 
Building on model predictive path integral (MPPI), we propose a cubic Hermite spline parameterization that operates on position and velocity control points. Our approach enables contact-making and contact-breaking strategies that adapt automatically to task requirements, requiring only a limited number of sampled trajectories. 
%
This sample efficiency enables real-time control on standard CPU hardware, eliminating the GPU acceleration typically required by other state-of-the-art MPPI methods.
We validate our approach on the Go2 quadrupedal robot, demonstrating a range of emergent gaits and basic jumping capabilities. In simulation, we further showcase more complex behaviors, such as backflips, dynamic handstand balancing and locomotion on a Humanoid, all without requiring reference tracking or offline pre-training. 
%
\end{abstract}

\section{Introduction}
\label{sec:introduction}
\begin{figure}[ht!]
    \centering
    \begin{subfigure}{0.9\linewidth}
        \centering
        \includegraphics[width=\linewidth]{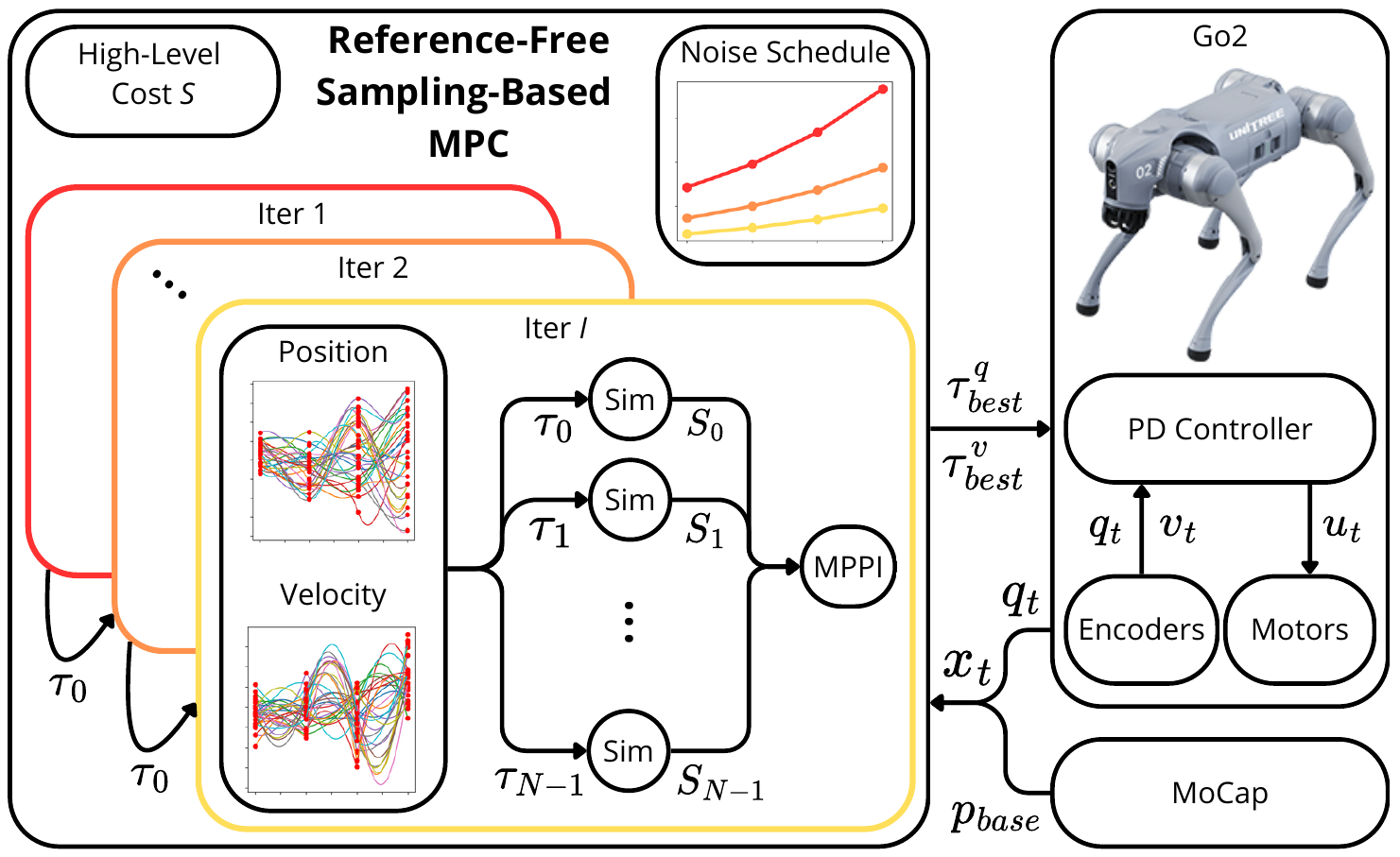}
        \caption{Overview of the framework showing the Hermite spline parametrization, noise schedule and reference-free costs.}
        \label{subfig:overview}
    \end{subfigure}
    \begin{subfigure}{0.9\linewidth}
        \centering
        \includegraphics[width=\linewidth]{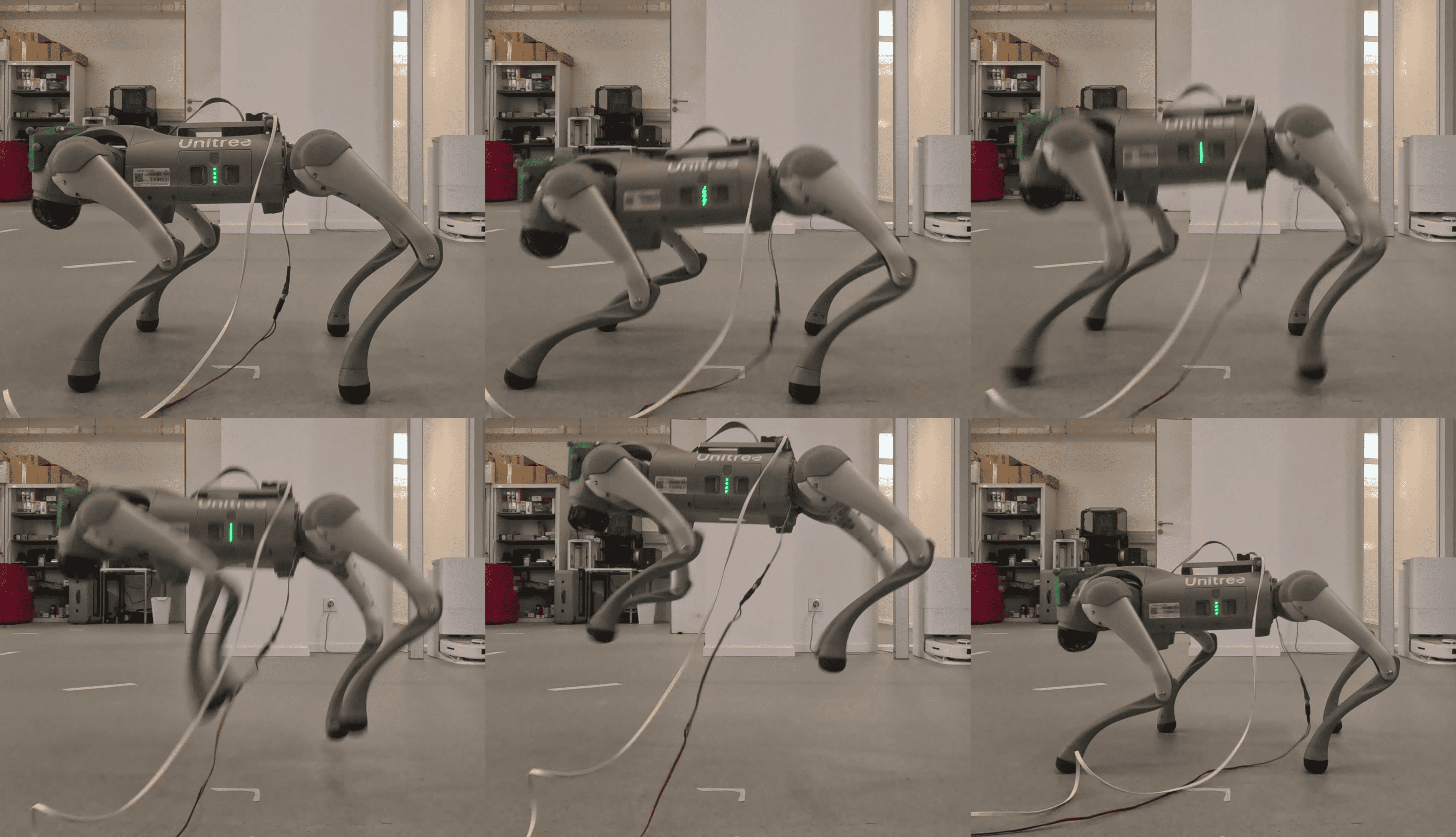}
        \caption{Jumping sequence where the robot crouches and leaps to a height of \SI{0.55}{\meter}.}
        \label{subfig:real-jump}
    \end{subfigure}
    \caption{Our reference-free sampling-based MPC framework enables emergent jumping motion experimentally achieved on the Go2 robot without any guiding reference.}
    \label{fig:go2-overall}
\vspace{-2em}
\end{figure}

Online robot control presents a trade-off between solution quality and computational efficiency. 
While learning-based methods such as reinforcement learning (RL) can generate impressive movements on complex robots, they require extensive offline training. They may also fail to generalize or adapt to new environments or tasks not seen during training~\cite{kober2013rlsurvey}. 
Despite ongoing efforts to improve sample efficiency~\cite{lidec2024diffsimple}, RL remains orders of magnitude too slow for online adaptation, and simplified models used to accelerate training can hinder sim-to-real transfer~\cite{deisenroth2011pilco, zhao2020sim2realreview}.
Typically, RL methods rely heavily on engineered rewards, such as phase clocks, air-time penalties, or foot-clearance objectives, to enforce structured contact behavior~\cite{hwangbo2019legged,Aractingi_2023,ha2025learning}.

On the other hand, trajectory optimization (TO) relies on derivative information to find high-quality solutions, but it requires accurate and informative gradients. This may be unavailable or unreliable in contact-rich scenarios, leading most TO methods to use predefined contact sequences~\cite{budhiraja2018differential,jalletPROXDDPProximalConstrained2025}.
Contact-implicit TO removes the need for predefined contact schedules~\cite{posa2014contactImplicit} and has enabled advanced demonstrations of complex whole-body behaviors~\cite{neunertWholeBodyNonlinearModel2018,kimContactimplicitModelPredictive2025}. However, many formulations rely on simplified contact models or approximations to remain tractable, which can introduce a mismatch with real dynamics and produce only approximate solutions in practice.
These methods also necessitate hand-crafted cost functions to obtain good contact sequences~\cite{kimContactimplicitModelPredictive2025}.

Sampling-based methods offer an attractive alternative by providing derivative-free optimization that is well-suited to parallel computation. This makes them particularly attractive for trajectory optimization problems with non-smooth dynamics. 
However, naive sampling approaches, such as random search, suffer from poor sampling efficiency and slow convergence, and may fail to converge to accurate solutions. 
Advances in sampling-based control over the past few years, particularly Model Predictive Path Integral~(MPPI) control~\cite{theodorou2017mppi}, despite their relative simplicity, have demonstrated promising results by incorporating more advanced sampling strategies and trajectory averaging schemes.

In this paper, we present a reference-free robotic control framework that challenges the paradigm requiring hand-crafted locomotion patterns. While recently introduced sampling-based MPC methods rely on gait references, whether through Raibert heuristics~\cite{alvarezpadilla2024realtimewholebodycontrollegged}, foot swing trajectories~\cite{Turrisi2024OnTB}, or predefined contact sequences~\cite{xue2024fullordersamplingbasedmpctorquelevel}, our approach enables the discovery of emergent locomotion purely from high-level goal specifications within a cost function.
In summary, the key contributions are:
\begin{itemize}
    \item A sampling-based MPC framework that enables reference-free motion discovery without reliance on gait priors or offline pre-training.
    \item A cubic Hermite spline parameterization that jointly samples position and velocity control points to improve exploration and dynamic consistency. Constraining endpoint derivatives allows for bound-preserving rules that prevent overshooting joint limits.
    \item Demonstration of real-time performance on standard CPU hardware with as few as 30 samples, validated both on a real platform and in a high-fidelity simulator. 
\end{itemize}

\section{Related Work}
\label{sec:related-work}
\begin{figure*}[ht!]
  \centering
  \includegraphics[width=\linewidth]{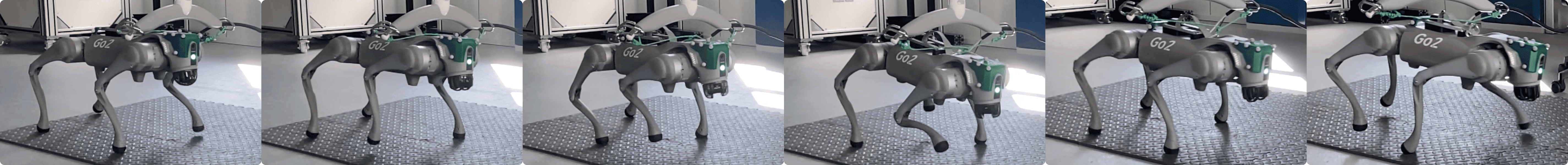}
  \caption{Sequence illustrating the discovered walking gait on the Go2 quadruped.}
  \label{fig:real-go2-walking}
\vspace{-1.5em}
\end{figure*}

\begin{table}[t]
\centering
\scriptsize
\rowcolors{2}{gray!30}{white}
\begin{tabular}{lccc}
\toprule
& DIAL-MPC & RT-Whole-Body & \textbf{Ours} \\
& \cite{xue2024fullordersamplingbasedmpctorquelevel} & MPPI \cite{alvarezpadilla2024realtimewholebodycontrollegged} & \\
\midrule
Samples & 2048 - 4096 & 30 & 30 - 70 \\
References & Swing foot & Raibert & \textbf{None} \\
Noise & Diffusion-inspired & Fixed & Diffusion-inspired \\
Hardware & GPU & CPU & CPU \\
Spline Type & Quadratic & Cubic & Hermite Cubic \\
Frequency & 50 Hz & 100 Hz & 50 Hz \\
Simulator & \parbox[t]{2cm}{MuJoCo MJX} & MuJoCo C++ & Simple \\
\bottomrule
\end{tabular}
\caption{Comparison of sampling-based MPC for legged robot control. We combine the noise-annealing strategy with the computational efficiency of prior work while eliminating the need for reference gait requirements.}
\label{tab:method-comparison-parbox}
\vspace{-2em}
\end{table}

Several works have explored sampling-based methods for robotic control. A common strategy is to refine a nominal action sequence iteratively, typically parameterized as a spline, using random perturbations.
%
Howell et al.~\cite{howell2022predictivesamplingrealtimebehaviour} evaluate predictive sampling (PS) as a zero-order baseline for comparison against methods like gradient descent, the iterative Linear Quadratic Regulator\,(iLQR)~\cite{Li2004IterativeLQ} and MPPI~\cite{theodorou2017mppi}. 
Despite its simplicity, PS achieves competitive performance and supports real-time tuning. The reported experiments, however, were restricted to torque-space splines in the \texttt{MuJoCo}~\cite{Todorov12mujoco} simulator and required workstation-class hardware.

More informative updates arise from leveraging statistics across all sampled trajectories, as in MPPI. This has shown strong performance in real-time control for racing, where parallel rollouts can be combined efficiently~\cite{williams2016aggressive_driving}. Related work has explored improving MPPI through structured proposal distributions, for example, via feature-based sampling~\cite{homburger2022featurebased}.
Turrisi et al.~\cite{Turrisi2024OnTB} demonstrated MPPI on a 12-DOF quadruped using GPU-accelerated rigid-body dynamics, achieving 10k rollouts for a 12-step horizon in under \SI{20}{\milli\second}, but with gait frequencies fixed a priori. 
Extensions to MPPI include DIAL-MPC~\cite{xue2024fullordersamplingbasedmpctorquelevel}, which introduces a two-level annealing schedule inspired by diffusion models~\cite{pan2024modelbaseddiffusiontrajectoryoptimization}. This method broadens exploration in early iterations and later horizon steps, while gradually refining actions closer to execution. Evaluation relies on GPU-based simulation with thousands of parallel rollouts and reference gait tracking. Building on DIAL-MPC, Crestaz et al.~ \cite{crestaz2025td} add a constraint-enforcing mechanism and a terminal value function approximation for longer-horizon reasoning.

Recent work has also investigated CPU-based implementations. Alvarez-Padilla et al.~\cite{alvarezpadilla2024realtimewholebodycontrollegged} propose real-time MPPI in joint space at \SI{100}{\hertz}, with torques generated via a PD controller at \SI{20}{\kilo\hertz}. Their system rolls out 30--50 trajectories in \texttt{MuJoCo}~\cite{Todorov12mujoco}, avoiding fixed contact sequences while still relying on reference heuristics for robust behavior.  

In contrast to these approaches, our method emphasizes low-sample efficiency and generality. It operates entirely on the CPU, uses a limited number of rollouts, does not rely on gait priors, and employs cubic Hermite splines to sample both position and velocity targets for the PD controller. Unlike prior joint-space sampling methods, which provide only position references (with velocity targets set to zero), our formulation yields dynamically consistent PD targets and enables richer exploration.
\section{Methodology}
\label{sec:methodology}

We formulate our controller within the MPPI framework~\cite{theodorou2017mppi}, which provides a sampling-based approach to trajectory optimization in a receding-horizon loop. At each control step, MPPI maintains a nominal control sequence over a finite horizon, perturbs this sequence with Gaussian noise, evaluates the resulting trajectories under a cost objective, and updates the nominal sequence via importance-weighted averaging before executing the first control input. 

Building on this foundation and using a diffusion-inspired annealing scheme that structures noise, we introduce several practical enhancements. We use a cubic Hermite spline parameterization that jointly optimizes position and velocity control points to improve exploration and enforce dynamically consistent trajectories, and a reference-free cost formulation that supports emergent locomotion without gait priors. To further improve stability and convergence, we integrate state prediction and warm-starting strategies to maintain temporal consistency across optimization steps. 
A complete overview of the framework is shown in Fig.~\ref{subfig:overview} and Tab.~\ref{tab:method-comparison-parbox} presents a comparison with recent MPPI-based methods.

\subsection{Search-space parametrization}

Defining the search space is critical for effective random-search optimization. In contrast to approaches that sample directly in torque space, often together with massively parallel GPU simulation~\cite{vlahov2024mppi}, we sample reference trajectories in joint space and track them using a PD controller that maps joint-space references to torque commands. This decouples high-level trajectory exploration from low-level torque regulation, enabling efficient CPU-based simulation at high frame rates. In our implementation, this requires only 20--30 parallel trajectories for quadruped control and 60--70 for humanoid control. \\[-1em]

\noindent\textbf{Cubic Hermite spline parameterization.}
For each actuated degree of freedom, the control sequence is parameterized by a cubic Hermite spline with $K$ nodes. The optimization variables are the node parameters
$$
\theta_k = (\theta_k^q, \theta_k^v), \qquad k = 0,\dots,K-1,
$$
where $\theta_k^q = q(t_k)$ and $\theta_k^v = \dot q(t_k)$ denote the joint position and joint velocity at node $k$. Throughout the paper, the superscript~${}^v$ denotes joint-velocity quantities associated with $q$.
%
%
The nodes are uniformly spaced in time with interval $\Delta t$, and we consider now one spline segment over $t \in [t_k, t_{k+1}]$. With the normalized local time
\begin{equation}
s = \frac{t - t_k}{\Delta t} \in [0,1],
\end{equation}
the joint trajectory is reconstructed using the standard cubic Hermite interpolant~\cite{deboor2001splines}
\begin{equation}
\begin{aligned}
q(t) &= h_{00}(s)\,\theta_k^q \;\;\:+ h_{10}(s)\,\Delta t\,\theta_k^v \\
&+ h_{01}(s)\,\theta_{k+1}^q + h_{11}(s)\,\Delta t\,\theta_{k+1}^v,
\end{aligned}
\label{eq:hermite_segment}
\end{equation}
\begin{align}
\text{where} \quad h_{00}(s) &= 2s^3 - 3s^2 + 1, & h_{10}(s) &= s^3 - 2s^2 + s, \\
h_{01}(s) &= -2s^3 + 3s^2, & h_{11}(s) &= s^3 - s^2
\end{align}
are the cubic Hermite basis functions. 
%
This parameterization is applied only to the actuated joints. The floating-base motion is not parameterized directly. Instead, it emerges from the forward dynamics rollout under the sampled joint references and contact interactions.

\begin{figure}[b]
    \centering
    \includegraphics[width=0.95\linewidth]{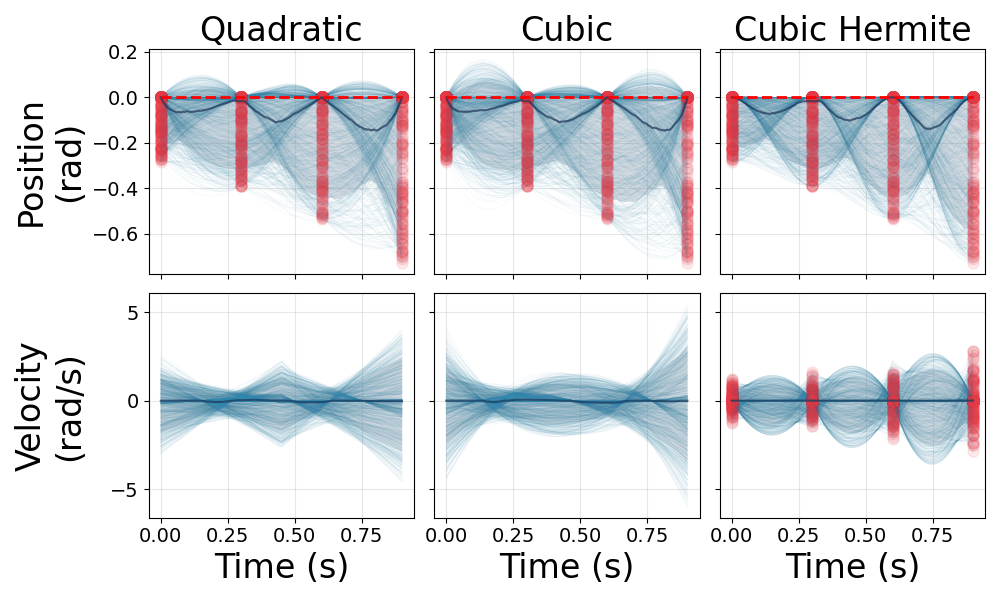}  
    \caption{Comparison of spline parameterizations for the same interpolation points (red). Quadratic, cubic, and cubic Hermite splines produce different position and velocity profiles. By parameterizing both node positions and velocities, cubic Hermite splines provide additional local shape control. Derivative clamping reduces overshoot near joint limits.}
    \label{fig:compare_splines}
\end{figure}
%
\noindent Compared with position-only spline parameterizations, cubic Hermite splines control both node positions and node velocities. This provides additional local control over the shape of the generated reference trajectories, as illustrated in Fig.~\ref{fig:compare_splines}.
To reduce overshoot near joint limits, we limit per-node derivatives based on the distance to the nearest bound. For node value $\theta_k^q \in [q_{min}, q_{max}]$ and node spacing $\Delta t$, we impose
\begin{equation}
     \label{Eqn:vel-clamp}
    |\theta_k^v| \le \frac{\min\{q_{max}-\theta_k^q,\, \theta_k^q-q_{min}\}}{\Delta t / 2} .
\end{equation}
This inexpensive clamp reduces mid-interval excursions and improves preservation of bounds in practice. In Fig.~\ref{fig:compare_splines}, the red dashed lines indicate the upper position bounds. While quadratic and cubic position splines often overshoot these bounds, our cubic Hermite formulation respects them throughout the trajectory. 
In the current formulation, the spline parameterization enforces position bounds via the derivative clamp above, but does not explicitly enforce joint-velocity limits at any intermediate time. In practice, feasibility is further shaped by the rollout dynamics, PD tracking, and the cost function.
%


\subsection{Noise annealing}  
The annealing schedule is motivated by the iterative nature of receding-horizon MPC and follows the diffusion-inspired design introduced in DIAL-MPC~\cite{xue2024fullordersamplingbasedmpctorquelevel}. A nominal control sequence is refined over $I$ internal iterations at each time step before executing the first action, after which the horizon advances. Consequently, control decisions farther in the horizon receive more updates (and thus can be explored more aggressively) while controls near execution should be more stable, with less variance.\\[-1em]

\noindent\textbf{Trajectory-level annealing.}
We therefore reduce the variance of exploration across iterations, transitioning from exploration to exploitation as the nominal trajectory is refined. For spline control points, this corresponds to shrinking the sampling covariance over iterations $i=I,\dots,1$:
\begin{equation}
    \det\!\big(\Sigma^i_{\theta}\big) \propto \exp\!\Big(-\frac{I-i}{\beta_1 I} K d_u\Big),
    \label{Eqn:traj-annealing}
\end{equation}
where $\beta_1$ is a temperature parameter, $K$ is the number of spline points, and $d_u$ is the per-node control dimension. \\[-1em]

\begin{figure}[t!]
  \centering
  \includegraphics[width=0.9\linewidth]{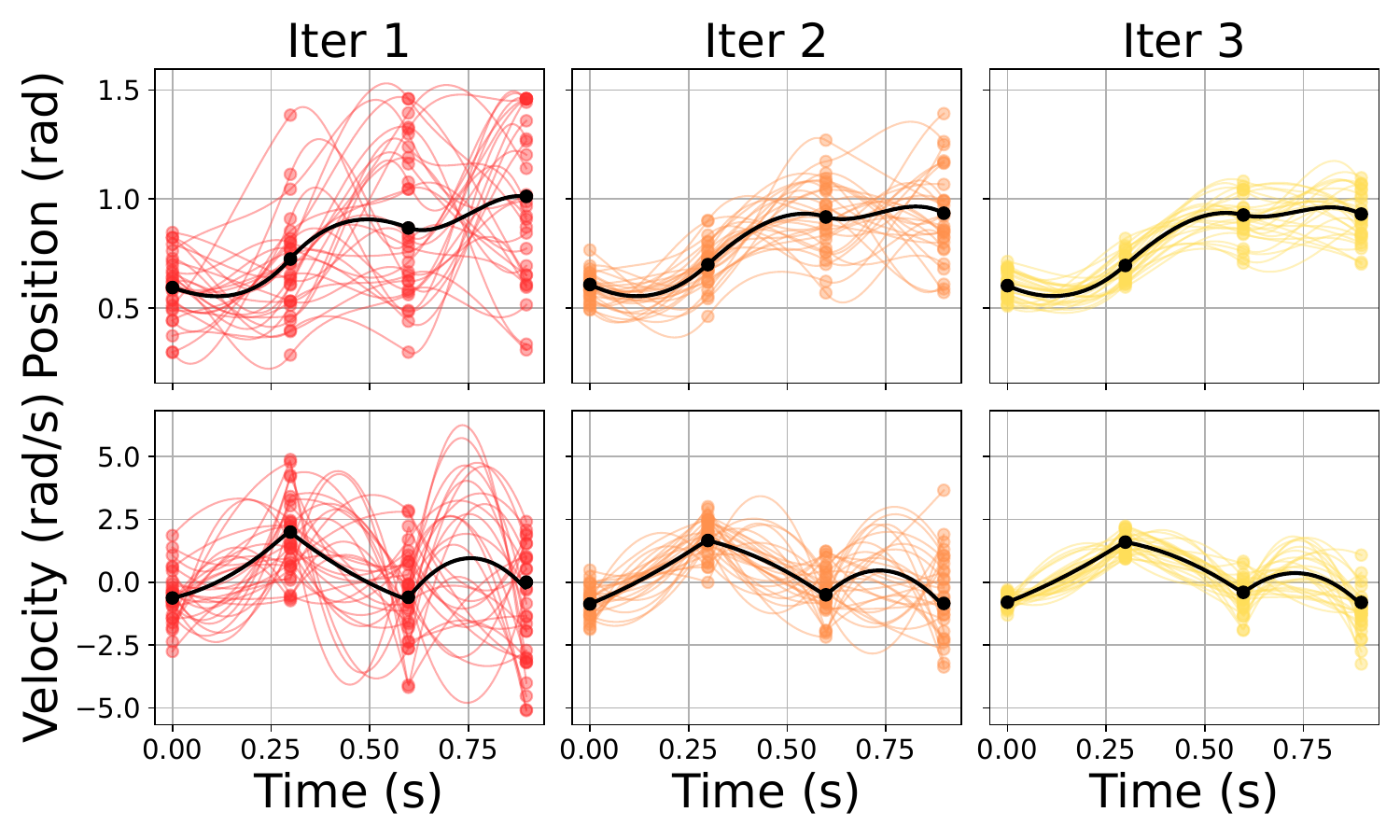}
  \caption{The nominal trajectory (black) evolves through spline control points that are updated iteratively. New perturbed spline points are sampled around the nominal points with annealing noise according to Eq.\,\eqref{Eqn:overall-noise-kernel}.}
  \label{fig:noise-schedule-joint-2}
  \vspace{-0.8em}
\end{figure}

\noindent\textbf{Action-level annealing.}  
We also increase exploration of control points corresponding to later actions in the horizon. For spline node index $k\in\{0,\dots,K-1\}$, this yields
\begin{equation}
    \det\!\big(\Sigma^i_{\theta_k}\big) \propto \exp\!\Big(-\frac{K-k}{\beta_2 K} d_u\Big),
    \label{Eqn:action-annealing}
\end{equation}
with $\beta_2$ controlling the horizon-wise decay. In our implementation, we use an isotropic, multiplicative combination of the two effects and parameterize the per-node covariance:
\begin{equation}
    \label{Eqn:overall-noise-kernel}
    \Sigma^i_{\theta_k} = \exp\!\Big(-\frac{I-i}{\beta_1 I} - \frac{K-k}{\beta_2 K}\Big)\mathbf{I},
\end{equation}
where $\mathbf{I}\in\mathbb{R}^{d_u\times d_u}$. The schedule in Eq.\,\eqref{Eqn:overall-noise-kernel} can be precomputed and cached once. 
It provides larger variance for later nodes and earlier iterations, and smaller variance for near-term nodes and later iterations. Fig.~\ref{fig:noise-schedule-joint-2} visualizes this behavior for a representative horizon and iteration count.

\subsection{Algorithm description}
The algorithm maintains a nominal sequence of spline control points in joint position $\theta^q$ and velocity $\theta^v$, defining a smooth trajectory over the prediction horizon via cubic Hermite interpolation. The position control points are initialized from a stable standing configuration and the velocity control points are initialized to zero.
At each control step, new perturbed control points $(\theta_n^q, \theta_n^v)$ are sampled around the nominal points with structured noise from the annealing scheme. Interpolating these perturbed points yields a batch of candidate trajectories $\tau_{n} = (\tau_{n}^q, \tau_{n}^v)$, which are rolled out in \texttt{Simple}~\cite{carpentier2024compliantrigidcontactsimulation,sathya2026constrained}, evaluated under the task cost, and assigned importance weights. The nominal sequence is then updated via importance-weighted averaging, and new control points are extracted by resampling the updated trajectory at spline node times, uniformly distributed across the horizon.

The exponential weighting scheme considers the relative quality of all candidates (see Eq.\,\eqref{Eqn:omega-weights}) and enables effective trajectory synthesis. The softmax smoothing concentrates probability mass around high-quality solutions and can be interpreted as performing approximate natural gradient descent on a smoothed surrogate function \cite{jordana2025introductionzeroorderoptimizationtechniques}.
The normalized weight $\omega_n$ for each trajectory $n$ is computed as
\begin{equation}
  \label{Eqn:omega-weights}
  \omega_n = \frac{\exp\!\big(-S_n/\lambda\big)}{\sum_{j}\exp\!\big(-S_j/\lambda\big)},
\end{equation}
where $S_n$ is the cost of trajectory $n$ and $\lambda$ is a temperature parameter that controls the selectiveness of the weighting.
In practice, we implement the calculation of $\omega_n$ using min-max-normalization of costs~\cite{JMLR:v11:theodorou10a}:
\begin{equation}
  \widehat S_n = \frac{S_n - S_{\min}}{S_{\max} - S_{\min}}, \quad 
    \,S_{\max} = \max_j S_j, \quad
    \,S_{\min} = \min_j S_j,
\end{equation}
and the weights are computed as
\begin{equation}
  \label{Eqn:softmax_minmax}
  \omega_n =\
  \frac{\exp\!\big(-\,\widehat S_n / \lambda\big)}
  {\displaystyle\sum_j \exp\!\big(-\,\widehat S_j / \lambda\big)}.
\end{equation}
The nominal control sequence $\tau_0$ is then updated as a weighted average of the candidates:
\begin{equation}
\label{Eqn:update-nominal}
\tau_{0}^q = \sum_{n} \omega_n \tau_n^q, \qquad
\tau_{0}^v = \sum_{n} \omega_n \tau_n^v.
\end{equation}
This update gradually shifts the nominal trajectory towards higher-quality solutions while retaining exploration diversity, and new nominal control points can be extracted.
The executed torque command is computed by a low-level PD controller from the computed best joint position and velocity targets to actuator torques
\begin{equation}
    \label{Eqn:pd-controller}
    u_t = K_p\big(\tau_{best}^q[0] - q_t\big) + K_d\big(\tau_{best}^v[0] - v_t\big),
\end{equation}
where $q_t,v_t$ are the measured joint positions and velocities, and $K_p,K_d$ are diagonal gain matrices. Torques are clipped to actuator limits before application.  

Overall, we build on MPPI with noise annealing and extend it with two key enhancements that enable reference-free locomotion discovery with a low number of samples:  
(1)~cubic Hermite spline sampling, which jointly perturbs position and velocity control points $(\theta^q, \theta^v)$ with physics-aware scaling, and  
(2)~best trajectory tracking, which separates the evolving nominal sequence $\tau_0$ from the executed actions $\tau_{best}$ for robustness and as a safeguard for consistent performance across iterations.
Alg.~\ref{algo:mppi} summarizes our complete approach, integrating these aspects within the MPPI framework.

\begin{algorithm}[t]
\caption{Reference-Free MPPI}
\label{algo:mppi}
\begin{algorithmic}[1]
    \Require parameters $H$, $K$, $I$, $N$, $\lambda$, $(\beta_1,\beta_2)$, ($K_p, K_d$)

    \State init. nominal spline control points $\theta^q \gets q_{0}, \theta^v \gets 0$
    \State $ \tau_{best} := \tau_{0} = (\tau_{0}^q, \tau_{0}^v) = \text{CubicHermite}(\{\theta^q\}, \{\theta^v\})$
    \State define annealed noise factors $\sigma^i_k$ using Eq.\,\eqref{Eqn:overall-noise-kernel}
    \For{every control step $t$}
        \State state prediction and warm-start (Sec. \ref{subsec:state-pred-warm-start})
    \State $\tau_{0} \gets \tau_{best}$ \Comment{init from best known trajectory}
    \For{each planning iteration $i \in \{I, \dots, 1\}$}
        \State extract nominal control points $\theta^q, \theta^v$ from $\tau_{0}$
        
        \For{each sample $n \in \{1, \dots, N-1\}$}
            \For{each spline point $k \in \{0, \dots, K-1\}$}
                \State sample $\theta_{n,k}^q \sim \mathcal{N}(\theta_{k}^q, \sigma^i_k \cdot \text{scale}_q)$
                \State sample $\theta_{n,k}^v \sim \mathcal{N}(\theta_{k}^v, \sigma^i_k \cdot \text{scale}_v)$
            \EndFor
        \State $\tau_n = (\tau_{n}^q, \tau_{n}^v) = \text{CubicHermite}(\{\theta_{n}^q\}, \{\theta_{n}^v\})$
        \EndFor

        \State simulate $\{\tau_n\}_{n=0}^{N-1}$ and compute costs $\{S_n\}_{n=0}^{N-1}$
        \State compute weights $\omega_n$ using Eq.\,\eqref{Eqn:softmax_minmax}
        
        \State update nominal $\tau_{0}$ using Eq.\,\eqref{Eqn:update-nominal}

        \State update $\tau_{best}$ (Sec. \ref{subsec:best-traj-track})
        \EndFor
    \State compute torque $u_t$ using Eq.\,\eqref{Eqn:pd-controller}
    \EndFor
\end{algorithmic}
\end{algorithm}

\subsection{Best trajectory tracking}
\label{subsec:best-traj-track}
One distinction is the separation between trajectory evolution and action execution. 
While the nominal trajectory $\tau_0$ evolves via standard MPPI importance-weighted averaging across iterations, the robot always executes actions from $\tau_{best}$, the best-performing trajectory tested in simulation rollout. 
This serves two purposes. First, it ensures that executed actions always come from a verified, fully-simulated trajectory rather than from an untested weighted mixture, providing safety guarantees. Second, it prevents performance degradation during iterative refinement by maintaining monotonic improvement in the quality of the executed solution. 

\subsection{Real-time state prediction and warm-starting strategy}
\label{subsec:state-pred-warm-start}

A practical challenge in real-time MPC is that both the sampling and optimization stages require time. 
In our setup, a full MPPI update with three iterations and 30 samples requires typically $\Delta t \approx 20-30$\,ms for a quadruped, during which the robot continues to execute the previously optimized trajectory. By the time the new solution is available, the robot has already moved on, so directly applying its first control would be inconsistent.  

To address this issue, we predict the state the robot will reach when the optimization is complete. Starting from the last known state $x_{t-\Delta t}$, we simulate forward using the prefix of the previously best trajectory $\tau_{\text{best}}$:  
\begin{equation}
    x_t = \text{simulate}\big(x_{t-\Delta t}, \tau_{\text{best}}[0:\lfloor \Delta t / dt \rfloor], \Delta t\big).
\end{equation}
This predicted state $x_t$ is then used as the starting point of the subsequent optimization instance.  
We then shift $\tau_\text{best}$ forward by the number of actions already executed during the computation delay, %
\begin{equation}
    \tau_{\text{best}}[h] \leftarrow \tau_{\text{best}}[h + \lfloor \Delta t / dt \rfloor], 
    \quad h \in [0, H - \lfloor \Delta t/dt \rfloor),
\end{equation}
and pad the tail by repeating the final action to preserve the horizon length. This maintains the solution continuity across receding-horizon steps and avoids re-optimizing from scratch. To improve convergence across control steps, we initialize the nominal trajectory $\tau_0$ with the shifted best trajectory from the previous optimization $\tau_{best}$. 

Our approach differs from prior work that handles computation delays by constraining the first $\lfloor\Delta t/dt\rfloor$ actions across all parallel rollouts to match the actions executed during computation. While conceptually simpler, this wastes computational budget by forcing identical initial actions across all sampled trajectories, reducing exploration diversity. Instead, our state prediction strategy optimizes from the predicted future state, allowing all samples to explore freely from the start of their planning horizons, thereby maximizing the effective use of the limited sample budget.

\section{Experiments}
\label{sec:experiments}

We evaluate our framework through real-world hardware experiments on the Go2 quadruped and complementary studies in simulation, including validation on the G1 humanoid. The experiments are designed to highlight three key aspects: (1)~the emergence of diverse locomotion strategies without reliance on gait references, (2)~real-time execution with minimal computational resources, and (3)~adaptability across platforms, tasks, and challenging behaviors. Demonstrations from both hardware and simulation are included in the accompanying video.
\vspace{-0.5em}

\subsection{Experimental setup}
\noindent\textbf{Real-world experimental hardware (Go2 quadruped only).}  
Experiments are conducted on the Unitree Go2, a \SI{16}{\kilo\gram} quadruped with 12 actuated joints. Joint states are measured from onboard encoders, while accurate global tracking is provided by a $300\,\text{Hz}$ motion capture system. The high-level MPC controller runs at $50\,\text{Hz}$ on a Mac Studio M3 equipped with 16 efficient cores and outputs desired joint positions and velocities, which are then tracked by a low-level PD control on the robot running at $12\,\text{kHz}$.\\[-1em]

\noindent\textbf{Experiments in simulation (Go2 quadruped and G1 humanoid).}  
To complement hardware experiments, we use the \texttt{Simple} physics simulator~\cite{carpentier2024compliantrigidcontactsimulation}, which provides accurate and efficient~\cite{sathya2026constrained} frictional contact dynamics simulation by proper handling of nonlinear complementarity contact models~\cite{brogliato:hal-01349852}.
Simulation experiments serve two purposes: (i)~showing behaviors that are unsafe or impractical to test on hardware, such as backflips, aggressive jumps, and high-speed locomotion up to \SI{2.0}{\meter\per\second}, and (ii)~demonstrating cross-platform generalization by validating our method on the Unitree G1 humanoid.\\[-1em]

\noindent\textbf{Cost function design.}  
All behaviors are driven by a modular cost formulation composed of running costs $c_t(x_t,u_t)$ and a terminal cost $c_T(x_H)$:
\begin{equation}
    S \;=\; \sum_{t=0}^{H-1} c_t(x_t,u_t) \;+\; c_T(x_H).
\end{equation}
The running cost combines residual terms on base motion, joint states, and contacts:
\begingroup\small
\begin{IEEEeqnarray}{rCl}
c_t(x_t,u_t) &&= w_{h}\,\lvert p_{base,z}-p_{des,z}\rvert \;+\; 
                 w_{orient}\,\|\log_3(R_{base}^T R_{des})\|_2^2 \nonumber\\
&&+\, w_{q}\,\|q-q_0\|_2^2 
   + w_{c,vel}\,\|v_{c}\|_1 
   + w_{c,force}\,\|f_{c}-f_0\|_1,
\end{IEEEeqnarray}
\endgroup
where $p_{base}$ and $R_{base}$ denote the base position and orientation, $q$ the joint angles, $v_{c}$ contact velocities, $f_{c}$ contact forces, and $q_0$, $f_0$ denote constant initial joint configurations and forces. Quantities with the subscript “des” indicate task-specific desired targets. The terminal cost encourages velocity tracking through base displacement:
\begin{equation}
c_T(x_H) \;=\; w_{H}\,\|p_{base}(x_H)-p_{target}\|_1,
\end{equation}
with $p_{target} = p_{base}(x_0) + \dot p_{des}\cdot H\cdot dt$.
Task-specific behaviors are obtained by adjusting the various weights while leaving the algorithmic framework unchanged. Tab.~\ref{tab:cost-weights} summarizes the weights used across experiments.

\begin{table}[t]
\rowcolors{2}{gray!30}{white}
\centering
\caption{Cost weights for different experiments.}
\label{tab:cost-weights}
\begin{tabular}{lcccccc}
\toprule
\textbf{Task} & $w_{h}$ & $w_{orient}$ & $w_{q}$ & $w_{c,vel}$ & $w_{c,force}$ & $w_H$ \\
\midrule
Walking   & 1e2 & 10   & 0.0 & 0.5 & 5e-2  & 2.5e3 \\
Jumping   & 1.0 & 0.5  & 0.3 & 1.0 & 5e-4  & 2e3   \\
Handstand & 50  & 10   & 0.3 & 1.0 & 5e-4  & 0.0   \\
Backflip  & 1.0 & 0.5  & 0.3 & 1.0 & 5e-4  & 2e3   \\
Bipedal   & 10  & 1.0  & 0.3 & 1.0 & 5e-4  & 2e2   \\
\bottomrule
\end{tabular}
\vspace{-1.5em}
\end{table}

\subsection{Motion discovery}
A central contribution of our work is the demonstration of reference-free motion discovery in real time across diverse tasks and robot morphologies. We evaluate the quadruped's ability to discover motion strategies purely from high-level goals and show that our framework generalizes to a humanoid robot, enabling walking without tuning gait parameters or other motion priors. For the more aggressive behaviors shown in simulation, we do not claim fully optimized or hardware-ready motions, but rather proof-of-concept behaviors obtained from the same generic framework.

\subsubsection{Velocity-adaptive gaits}
\begin{figure}[t]
    \centering
    \begin{subfigure}{0.8\linewidth}
        \centering
        \includegraphics[width=\linewidth]{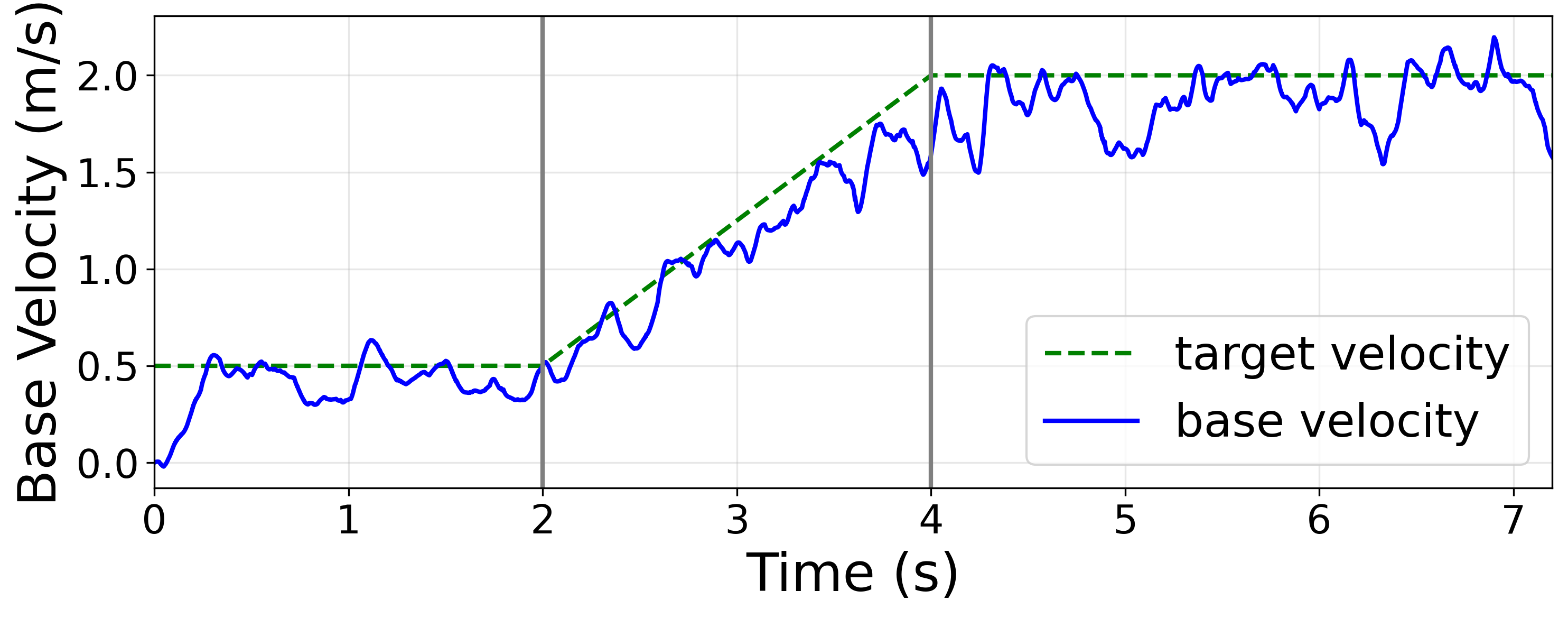}
        \caption{Base velocity tracking for different speeds.}
        \label{subfig:velocity-tracking}
    \end{subfigure}
    \begin{subfigure}{0.8\linewidth}
        \centering
        \includegraphics[width=\linewidth]{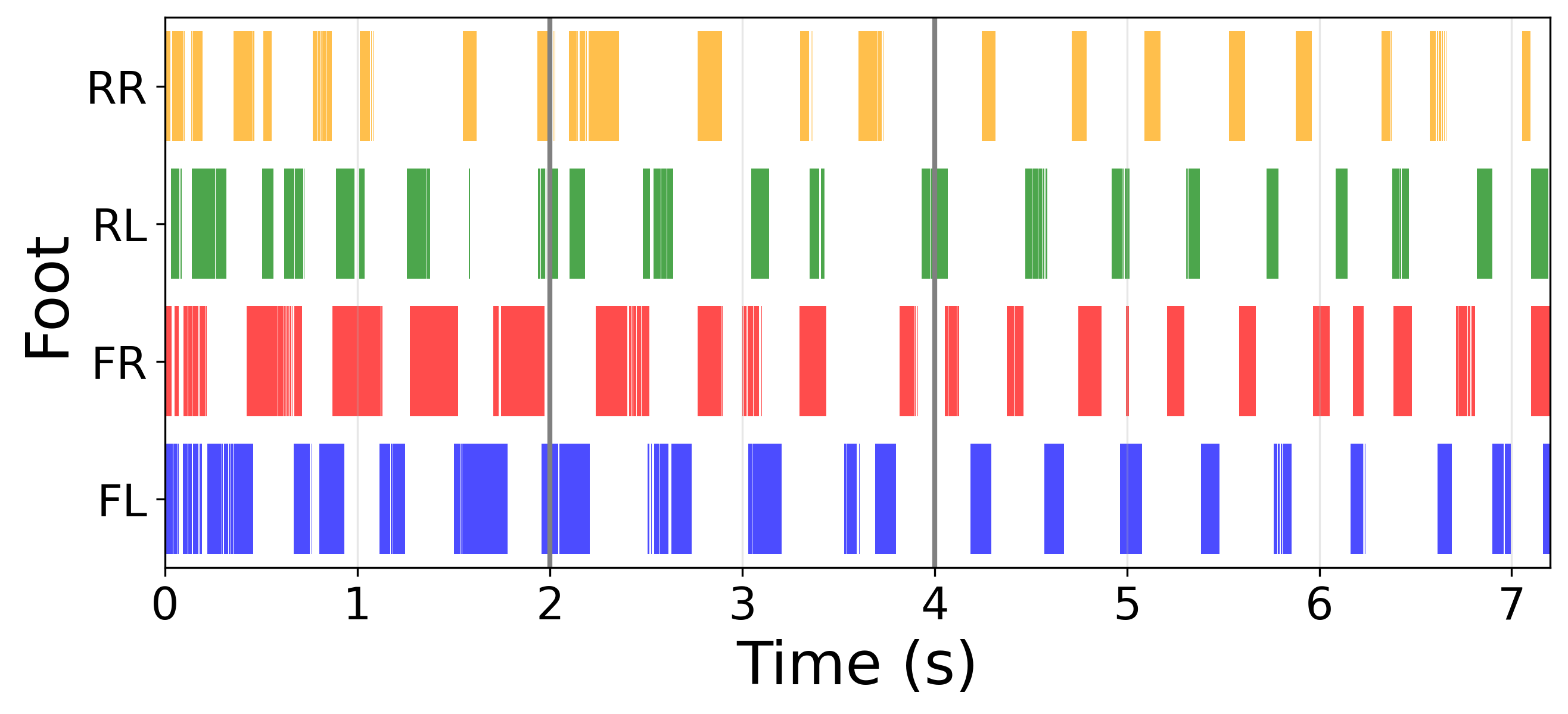}
        \caption{Contact pattern of the emergent gaits.}
        \label{subfig:contact-pattern}
    \end{subfigure}
    \caption{Smooth transitioning from trotting to galloping as velocity commands change, showing adaptive gait discovery.}
    \label{fig:trotting_gait}
    \vspace{-0.5em}
\end{figure}

In simulation and on the real Go2 robot, we specify the desired forward velocity in the cost function without providing any explicit gait references or contact sequence. 
We use an extended planning horizon of \SI{0.9}{\second} that is important for locomotion discovery. In contrast to reference-tracking methods that can use shorter horizons of \SI{0.4}{\second} to follow predefined patterns, emergent gait discovery requires sufficient horizon time to evaluate locomotion stability and quality. For slow walking in particular, a \SI{0.9}{\second} window allows the optimization to assess whether a motion pattern leads to stable periodic locomotion or to a fall.
We evaluate three representative velocity commands:

\begin{itemize}
    \item Standing still (\SI{0}{\meter\per\second}): The robot consistently adopts a stable posture with the 4 feet in contact, maintaining balance and rejecting disturbances through leg adjustments and stepping. 
    \item Trotting (\SI{0.5}{\meter\per\second}): A trotting gait emerges, characterized by alternating pairs of legs in contact.
    \item Galloping (\SI{2.0}{\meter\per\second}): The robot transitions to a more dynamic gait with extended flight phases, resembling a galloping or bounding pattern. This transition occurs smoothly as the velocity command increases.
\end{itemize}
To analyze the different gaits, Fig.~\ref{subfig:velocity-tracking} shows base-velocity tracking as the commanded speed is ramped from \SI{0.5}{\meter\per\second} to \SI{2.0}{\meter\per\second}. Fig.~\ref{subfig:contact-pattern} shows the contact patterns, highlighting the transition from trotting to more dynamic gaits as the velocity command increases. The resulting behavior on the real Go2 platform is illustrated in Fig.~\ref{fig:real-go2-walking}.

\subsubsection{Robustness}
To evaluate robustness, we conduct real-world disturbance tests over \SI{10}{\minute} of continuous operation. We command the robot to maintain a target position ($x$, $y$, $z$) and yaw angle while systematically applying various disturbances. We begin with direct physical perturbations that push the robot from multiple directions (sides, top, diagonally) and manually pull individual legs. The controller resists these external perturbations and generates corrective stepping motions without falling.

When we rotate the mattress on which the robot is standing, it responds by generating turning walking motions to restore its original yaw angle. If the mattress is pushed or pulled, the robot walks to recover its original position. In the most extreme test, we lift the robot into the air, rotate it by more than \SI{90}{\deg}, and set it back down; the robot then walks back to its target pose. These tests are illustrated in the companion video.

\subsubsection{Jumping}
We demonstrate diverse jumping behaviors by specifying only sparse task-level objectives where the goal is to jump vertically, turn the base, and land safely. \\[-0.5em]

\noindent\textbf{Vertical jumping}: Commanding a target base height increase (\SI{0.325}{\meter} → \SI{0.7}{\meter}) as terminal cost via a step function, the robot discovers a complete jumping strategy.
\begin{figure}[t]
    \centering
    \includegraphics[width=0.8\linewidth]{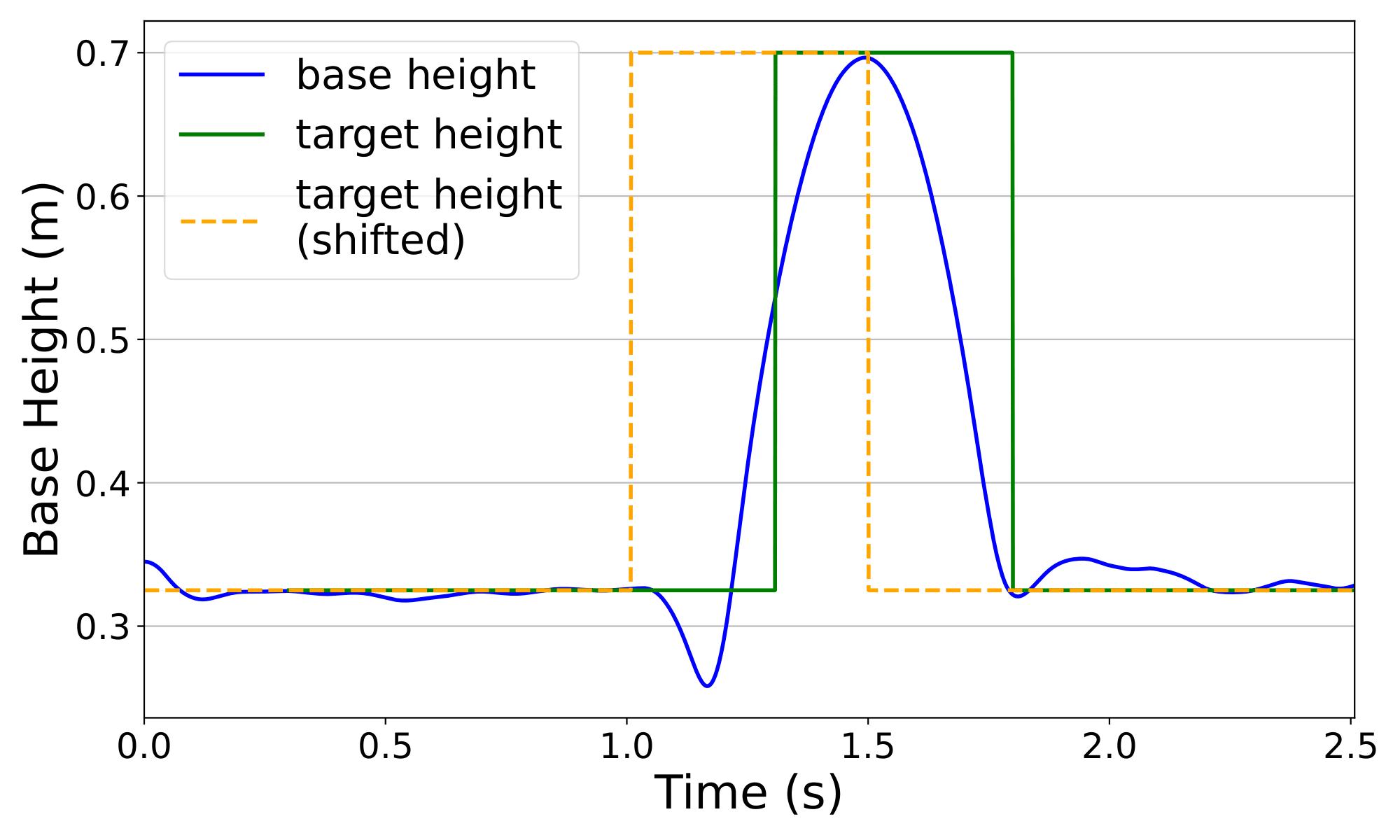}
    \caption{Robot base height during vertical jumping. When commanded to reach \SI{0.7}{\meter} (green), the robot discovers a jumping strategy. The orange dashed line shows when the terminal height objective enters the MPC horizon, triggering the robot to initiate its jump sequence (blue).}
    \label{fig:jump}
    \vspace{-1.5em}
\end{figure}
As soon as the terminal height objective becomes visible in the current MPC horizon (see orange dashed line in Fig.~\ref{fig:jump}), the robot initiates a multi-phase jumping maneuver. First, it crouches by flexing its legs, then rapidly extends them to launch into the air, reaching a peak height of \SI{0.7}{\meter} above the ground. During flight, the robot controls its body orientation to maintain stability and prepare for landing. 
We have successfully validated this jumping behavior on the real Go2 robot platform, as shown in Fig.~\ref{subfig:real-jump}. \\[-0.5em]

\begin{figure}[t]
  \centering
  \begin{subfigure}[b]{0.6\linewidth}
    \centering
    \includegraphics[width=0.95\linewidth]{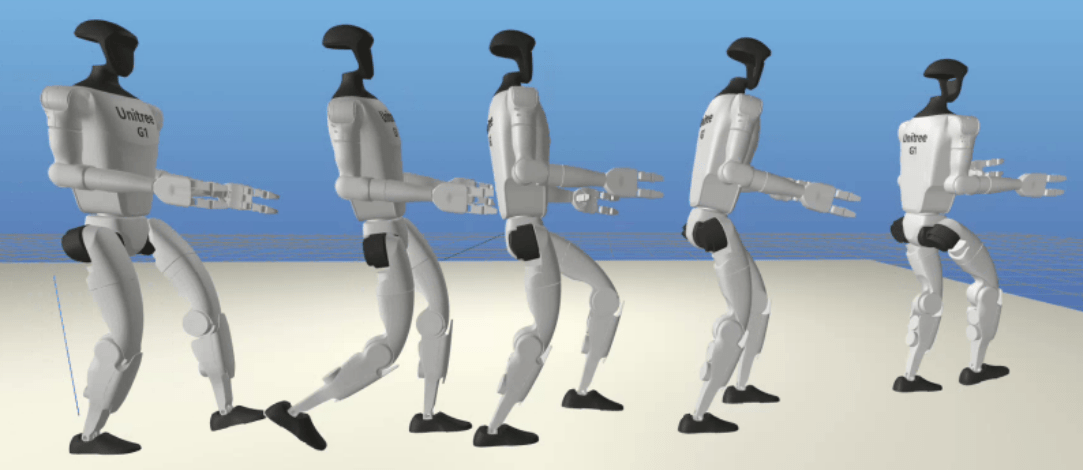}
    \caption{Humanoid locomotion.}
    \label{subfig:g1-walk}
  \end{subfigure}
  \hfill
  \begin{subfigure}[b]{0.35\linewidth}
    \includegraphics[width=\linewidth]{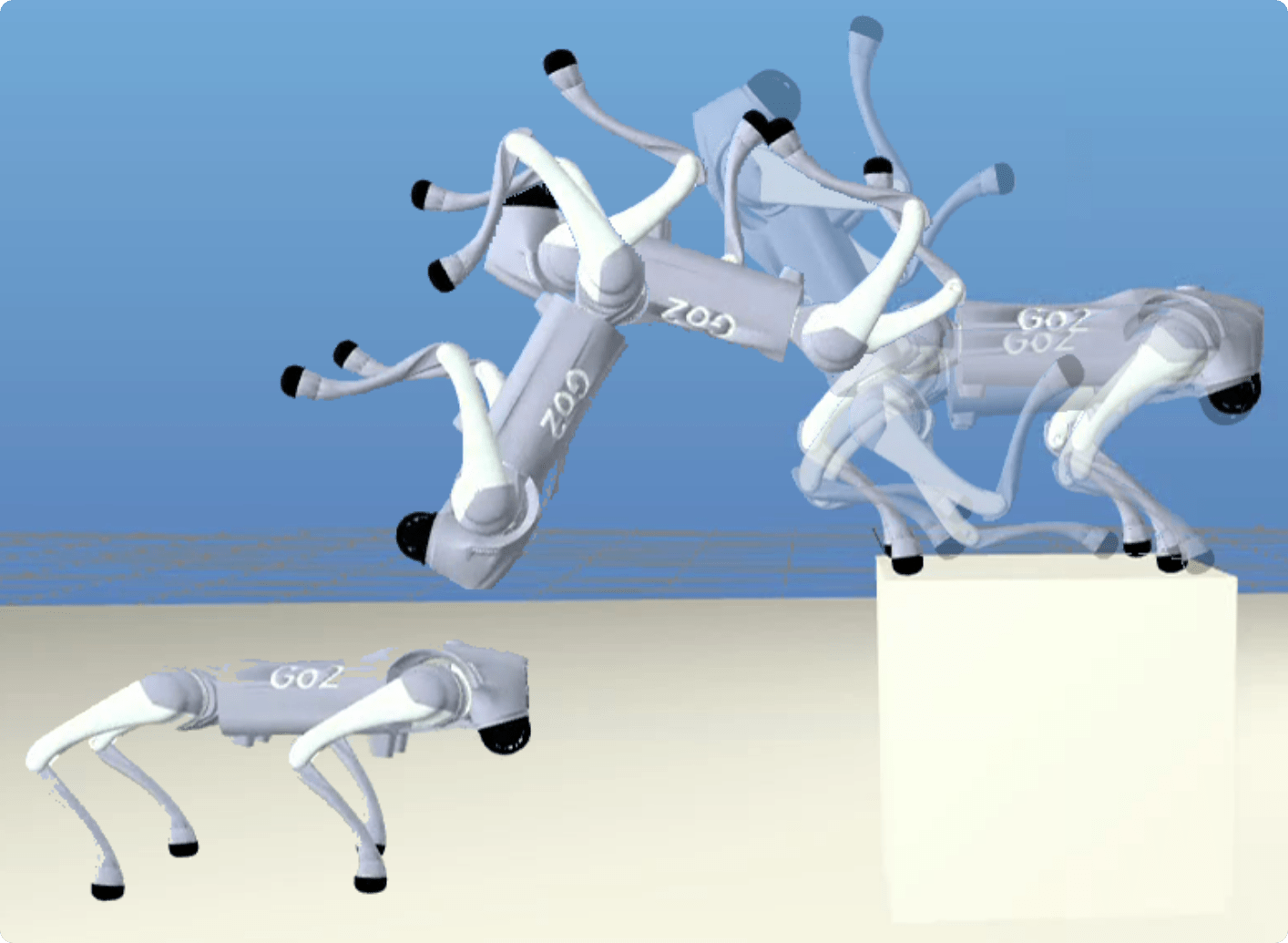}
    \caption{Backflip maneuver.}
    \label{subfig:backflip}
  \end{subfigure}
  \caption{Emergent dynamic behaviors in simulation.}
  \label{fig:sim-handstand-g1}
\vspace{-1.2em}
\end{figure}

\noindent\textbf{Backflip}: By specifying a target pitch orientation (\SI{180}{\degree}) at the highest point, the robot discovers a complete backflip maneuver. It coordinates a pre-jump crouch, an explosive takeoff that generates both vertical impulse and angular momentum, mid-air rotation control, and precise landing absorption to achieve the desired orientation and height. Starting from the elevated platform of \SI{0.5}{\meter}, the robot completes a full 360° rotation with landing as shown in Fig.~\ref{subfig:backflip}. 

\begin{figure}[t]
    \centering
    \begin{subfigure}{0.9\linewidth}
        \centering
        \includegraphics[width=0.95\linewidth]{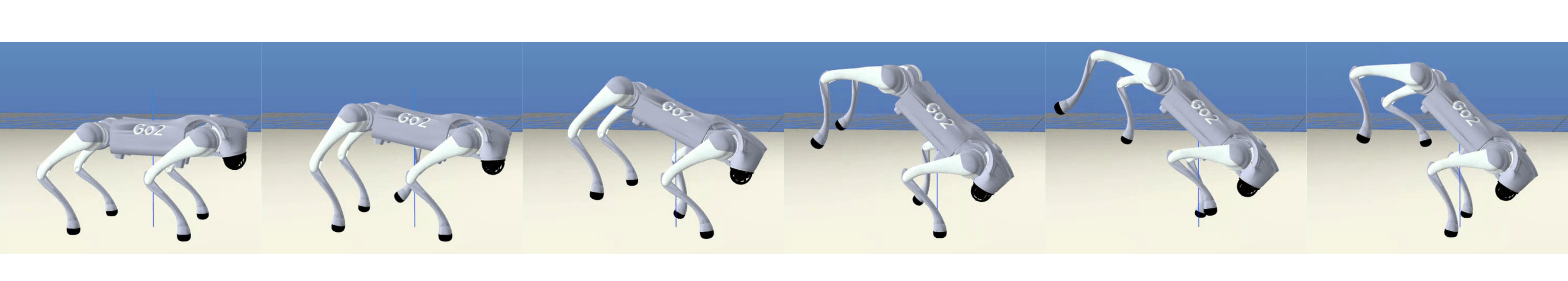}
        \label{subfig:handstand_series}
    \end{subfigure}
    \begin{subfigure}{0.9\linewidth}
    \includegraphics[width=0.95\linewidth]{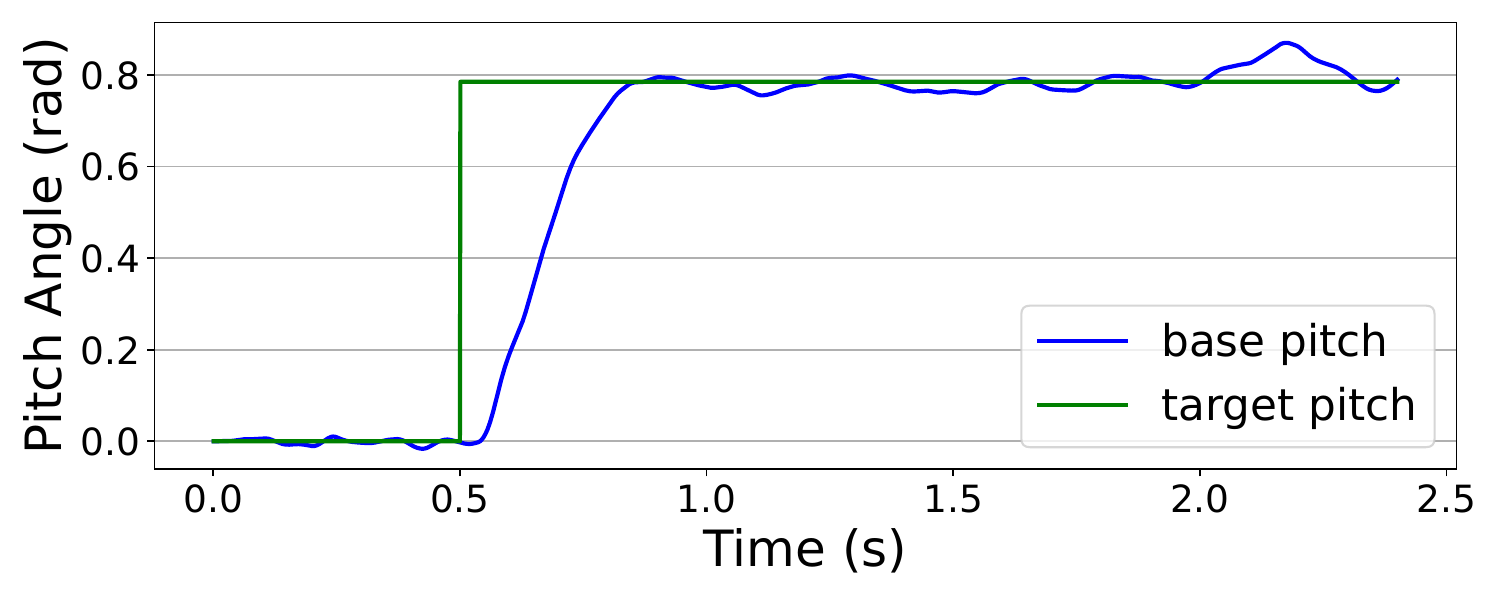}
        \label{subfig:handstand_plots}
    \end{subfigure}
    \caption{Base pitch trajectory during handstand pose.}
    \label{fig:handstand}
\end{figure}
\subsubsection{Dynamic handstand balancing}
Starting from a standard quadrupedal stance with a level base, when commanded to achieve an inverted pose (\SI{45}{\degree} pitch angle), the robot executes a dynamic swing maneuver to transition into a handstand configuration. Then, the controller identifies stabilizing strategies by continuously repositioning the legs and adjusting contact patterns. 
Fig.~\ref{fig:handstand} illustrates simulation snapshots of emergent contact-making and breaking patterns, where the legs dynamically reposition to maintain balance. The accompanying plot shows that the handstand pose is stably maintained for about \SI{1}{\second} with low orientation error.

\subsubsection{Humanoid locomotion}

To highlight the generality of our reference-free framework beyond quadrupedal locomotion, we also evaluate it on the G1 humanoid in simulation. Despite the shift in morphology and increase in DoFs (from 12 to 37), the same algorithm successfully discovers a walking gait from the exact high-level cost without modification, as displayed with overlayed snapshots in Fig.~\ref{subfig:g1-walk}.
%
%
While all previously demonstrated behaviors were achieved in real-time, the humanoid experiment requires an increase from 30 to 70 parallel sample trajectories due to the increased complexity of bipedal balance control, which prevents real-time execution.
This cross-platform study underlines a key advantage of our approach. The identical framework, cost functions, and spline parameterization transfer across robot morphologies, enabling rapid deployment on new platforms. However, these humanoid results are currently limited to simulation and should be interpreted as a proof of concept for motion discovery rather than as hardware-ready whole-body behaviors. The discovered motions are not yet optimized for smoothness, efficiency, or robustness to real-world sensing and actuation constraints.

\subsection{Ablation studies}

To better understand the contribution of our algorithmic components, we perform ablation studies that isolate and evaluate individual design choices. Specifically, we investigate two central enhancements of our framework:\\[-0.5em]

\noindent\textbf{(1) Cubic Hermite splines.}  We compare our cubic Hermite spline parameterization with velocity targets against cubic splines without velocity targets and quadratic splines. The goal is to quantify the improvement in trajectory quality and stability achieved by explicitly controlling both position and velocity at spline nodes. Note that we are using the best trajectory tracking mechanism~\ref{subsec:best-traj-track} in all cases. \\[-0.5em]

\noindent\textbf{(2) Best trajectory tracking.} We study the effect of executing actions from the best sampled trajectory $\tau_{best}$ rather than from the evolving nominal trajectory $\tau_0$. This ablation tests whether explicitly safeguarding execution against untested mixtures yields more reliable performance. In both cases, trajectories are parameterized using Hermite splines. We then track how often the optimizer failed to improve upon the shifted previous solution: walking $(27.9\%)$, jumping $(29.7\%)$, and handstand $(27.1\%)$.\\[-0.5em]

\begin{figure}[t]
    \centering
    \includegraphics[width=0.9\linewidth]{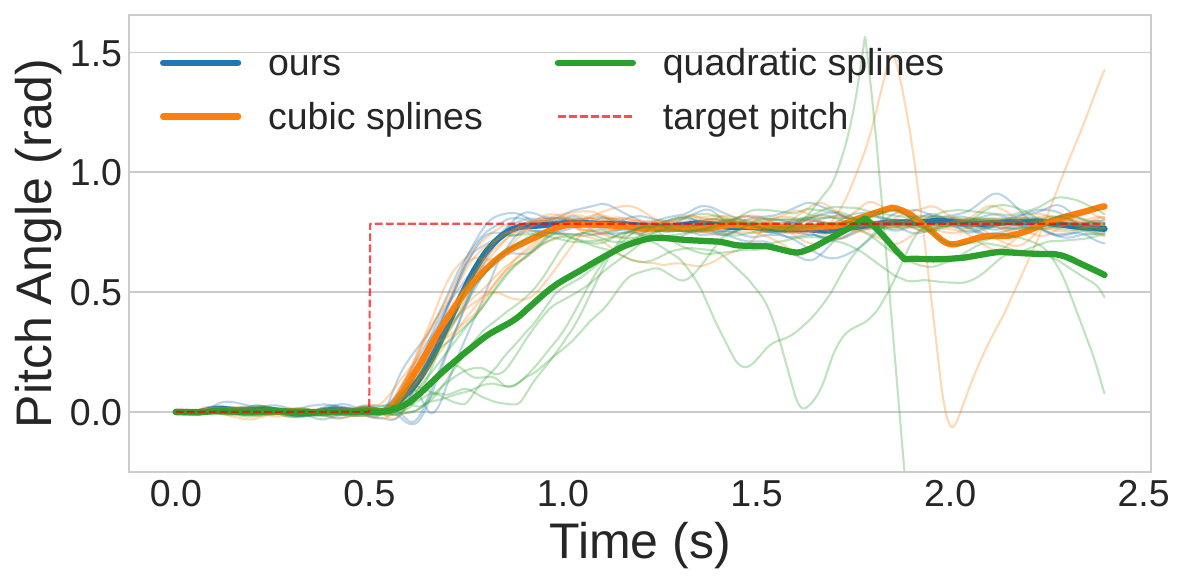}
    \caption{Handstand ablation: base pitch trajectories for different spline parameterizations during the handstand task.}
    \label{fig:handstand-ablation}
\vspace{-2em}
\end{figure}

For both ablations, we evaluate performance across multiple tasks, walking, handstand, and jumping, using 10 randomized seeds per setting. The primary metric is the average trajectory cost, which directly reflects the optimization objective. In Tab.~\ref{tab:ablations}, we report mean $\pm$ standard deviation to highlight consistency across runs. Failures are recorded when the robot base hits the ground. For the handstand task, Fig.~\ref{fig:handstand-ablation} additionally shows the base pitch trajectories. For the jumping task, we report the mean $\pm$ standard deviation of the maximum reached height to provide a more interpretable performance metric.
Across all tasks, our Hermite spline parameterization, combined with best-trajectory tracking, yields lower mean trajectory costs and lower variance. This advantage is particularly evident in jumping, where our method achieves the highest mean height (h), and in handstands, where cubic and quadratic splines exhibit larger variance and occasional failures.
%

\begin{table}[t]
\centering
\caption{Ablation results. Values are success rates (over 10 seeds) and mean $\pm$ std for continuous metrics.}
\label{tab:ablations}
\scriptsize
\setlength{\tabcolsep}{3pt}
\renewcommand{\arraystretch}{0.95}
\rowcolors{2}{gray!30}{white}
\begin{tabular}{lcccc|c}
\toprule
\rowcolor{white}
\multirow{2}{*}{\textbf{Variant}} & \multicolumn{2}{c}{Walk} & \multicolumn{2}{c}{Handstand} & \makecell{Jump} \\
\cmidrule(lr){2-3} \cmidrule(lr){4-5} \cmidrule(lr){6-6}
\rowcolor{white}
& Succ. & Cost & Succ. & Cost & \makecell{Max h.\\(cm)} \\
\midrule
Hermite (ours) & 10/10 & 913.9 $\pm$ 25.8   & 10/10 & ~\,811.7 $\pm$ 253.4  & 68.4 $\pm$ 0.2 \\
Cubic          & 6/10  & 1124.1 $\pm$ 135.7 & 10/10 & 1540.2 $\pm$ 655.8 & 63.8 $\pm$ 0.4 \\
Quadratic      & 0/10  & ~\,--                 & 9/10  & 1882.2 $\pm$ 493.9 & 47.7 $\pm$ 0.4 \\
\midrule
Best traj. (ours)    & 10/10 & 913.9 $\pm$ 25.8   & 10/10 & ~\,811.7 $\pm$ 253.4  & 68.4 $\pm$ 0.2 \\
Nominal only   & 10/10 & 924.9 $\pm$ 24.3   & 10/10 & ~\,844.1 $\pm$ 203.2  & 67.4 $\pm$ 0.3 \\
\bottomrule
\end{tabular}
\vspace{-1em}
\end{table}

\section{Discussion and Conclusion}
\label{sec:conclusion}

We presented a sampling-based MPC framework that enables reference-free locomotion by combining cubic Hermite spline parameterization with diffusion-inspired noise annealing. In contrast to prior MPPI approaches that rely on thousands of GPU rollouts and gait priors, our method achieves real-time performance on a CPU with as few as 30 samples.  
%
Experiments on the Go2 quadruped and G1 humanoid show that the same framework can produce diverse locomotion behaviors across tasks and robot morphologies without reference tracking or predefined contact sequences.

A limitation of the current formulation is that the resulting motions are not always fully optimized with respect to smoothness or efficiency. This follows from our design choice to sample spline control points in joint space and delegate torque-level execution to a PD controller, which reduces the search space but also limits fine-grained optimization compared with direct torque optimization.

This is a drawback of the present formulation, but also a trade-off that makes the approach useful as a motion-discovery mechanism. The framework provides a flexible source of candidate behaviors, contact sequences, and strategies that can serve as warm starts for higher-level solvers or be integrated into larger control pipelines. Future directions include improving the noise sampling scheme and integrating our method with trajectory optimization or learning-based refinement to transform diverse exploratory motions into polished, task-specific controllers.

\vspace{-0.1em}



\section*{Acknowledgments}
Supported by the French government via ANR under France 2030 (PEPR O2R, PR[AI]RIE-PSAI ANR-23-IACL-0008, RODEO ANR-24-CE23-5886); the EU (ARTIFACT 101165695, AGIMUS 101070165); and Région Île-de-France (DIM AI4IDF). 
%
Views and opinions expressed are those of the author(s) only and do not necessarily reflect those of the funding agencies.


\bibliographystyle{IEEEtran}
\bibliography{references}

\end{document}